# Variational Learning in Mixed-State Dynamic Graphical Models

**Vladimir Pavlovic**
Cambridge Research Lab
Compaq Computer Corp.
Cambridge, Massachusetts

**Brendan J. Frey**
Computer Science
University of Waterloo
Waterloo, Ontario

**Thomas S. Huang**
Beckman Institute
University of Illinois
Urbana, Illinois

## Abstract

Many real-valued stochastic time-series are locally linear (Gaussian), but globally non-linear. For example, the trajectory of a human hand gesture can be viewed as a linear dynamic system driven by a nonlinear dynamic system that represents muscle actions. We present a mixed-state dynamic graphical model in which a hidden Markov model drives a linear dynamic system. This combination allows us to model both the discrete and continuous causes of trajectories such as human gestures. The number of computations needed for exact inference is exponential in the sequence length, so we derive an approximate variational inference technique that can also be used to learn the parameters of the discrete and continuous models. We show how the mixed-state model and the variational technique can be used to classify human hand gestures made with a computer mouse.

## 1 Introduction

When reasoning about the causes of a temporal sequence of real-valued noisy observations, it is natural to postulate the existence of both continuous causes and discrete causes and that in general these causes are related probabilistically through time by a Markovian structure. For example, suppose we observe an autonomous moving target. The target motion is governed by Newtonian physics as well as the input force (thrust) imposed upon it by its human operator. If we have some knowledge of what sequences of actions the operator may take in time, it is natural to combine a real-valued dynamical model of the physics with a discrete-valued model of the input forces, where the number of discrete states corresponds to the number of possible actions the operator may take. Another example of a physical system that can be modeled in this way is the human hand/arm motion during gestural communication. The physical arm motion can be described using different kinematic and dynamic models of simple or articulated structures while the forces that influence the arm motion can be modeled using discrete hidden variables.

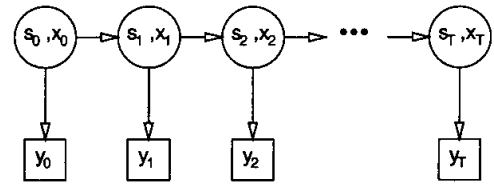

Figure 1: An unwrapped dynamic Bayesian network that uses both real-valued states x and discrete-valued states s to explain the observations y.

Although efficient algorithms exist for probabilistic inference in real-valued Gaussian chains (Kalman filtering [16]) and discrete hidden Markov models (the forward-backward algorithm [22]), combining the state spaces to produce a "mixed-state" dynamic graphical model with both real-valued Gaussian variables and discrete-valued variables makes exact probabilistic inference an exponentially difficult problem. From Fig. 1, it is clear that the distribution over the sequence of $T$ real-valued states $x_1, \ldots, x_T$ given the sequence of observations $y_1, \ldots, y_T$ *and* the sequence of discrete-valued states $s_1, \ldots, s_T$ will be described by a Gaussian chain. So, if each discrete state variable can take on $S$ values, the posterior distribution over the combined state sequence will be a mixture of $S^T$ Gaussian chains.

A fast, greedy way to approximate inference in mixed-state dynamic models is to simply apply the Viterbi algorithm in the Gaussian chain (Kalman filtering) while ignoring the discrete variables and then fix the resulting real-valued sequence while performing the Viterbi algorithm (or forward-backward algorithm) in the discrete chain. When there is little noise in the observations and the HMM dynamics emphasize discrete-valued sequences with distinct real-valued states, this method tends to work well, since the observations effectively determine the real-valued states. However, when there is a significant amount of noise in the observations or the discrete-valued state sequences are not distinct enough, this method fails to tailor the estimate of the real-valued states to the probable configurations of the discrete states.

Another approach to speeding up inference at the cost of approximating it is to consider how the exact inference algorithm (the forward backward algorithm) can be modified to avoid the exponential growth in the number of mixture components with time. A common method is to simply retain a fixed number of the most massive components. One extreme of this method results in another Viterbi-like algorithm, when only the most massive component is retained. Alternatively, the probability messages passed forward and backward in the chain can be approximated with a statistical model [17]. Boyen and Koller show in [3] that under some conditions, these types of approximation to the forward-backward algorithm will not lead to an accumulation of error with sequence length.

The Markov blanket property of graphical models often makes them suitable for Markov chain Monte Carlo methods. Given the Markov blanket of a variable, it is often easy to draw a sample for the variable. By repeating this procedure for latent variables chosen at random or in order, a sequence of states is obtained. The stationary distribution of this sequence of states is the posterior distribution[21, 18].

A different approach is to keep an inverse recognition model that can map the observations to an estimate of the hidden state distribution [6, 14, 8]. In fact, this recognition model can also be a graphical model that can represent covariance in the posterior distribution. The recognition model can be learned using the "wake-sleep" algorithm, which employs Monte Carlo sampling from the generative model to train the recognition model.

One disadvantage of all of the above methods is that the algorithms do not consistently decrease the relative entropy between the approximating distribution and the true posterior distribution. Even if a local approximation to the forward-backward algorithm does not lead to accumulating errors, it may be that incurring a larger error at one point in the chain would significantly reduce the errors at other points in the chain. Monte Carlo methods are only exact in the case of infinite size samples. For finite samples, they are not guaranteed to give a good approximation. Ensembles of samplers can be used to improve performance, but even so they are guaranteed to consistently improve the approximation to the true posterior only as the number of samplers tends to infinity. Although probabilistic recognition networks have the potential to approximate inference arbitrarily well, the "wake-sleep" algorithm used to train recognition networks decreases an *approximation* to the relative entropy.

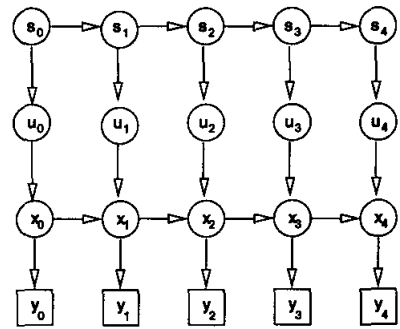

Figure 2: A mixed-state dynamic Bayesian network (DBN). $s_t$ is the state of the discrete-state hidden Markov model (HMM) at time $t$; $u_t$ is the output of the HMM at time $t$; $x_t$ is the state of the linear dynamic system at time $t$; and $y_t$ is the output of the mixed-state model at time $t$.

"Structured variational inference" [15] is a technique where the structure of the approximating distribution is explicit and the algorithm consistently decreases the distance between the approximating distribution and the true posterior distribution. In this paper, we present a structured variational inference method for inference and learning in mixed-state dynamic graphical models. In contrast to variational techniques in general dynamic networks with factored representations [12], our inference method takes advantage of the two computationally tractable substructures. Fig. 2 shows how a mixed-state dynamic model can be represented by a hidden Markov model (HMM) driving a linear dynamic system (LDS).

In related work, Ghahramani and Hinton [11] considered a different combination of discrete and real-valued variables. In their "switching state space model", an HMM chooses which of several LDSs will produce the current output. In our model, the HMM *drives* the LDS. Blom and Bar-Shalom [2] and West and Harrison [26] considered multiple LDS models whose activity is governed by Markovian dynamics. To maintain tractability of inference they utilized truncation of the mixture components. Other related work includes coupled HMMs [25, 4, 20] and mixtures of dynamic graphical models [20].

## 2   Mixed-state dynamic Bayesian networks

The mixed-state dynamical model can be described using the following set of state-space equations:

$$x_{t+1} = Ax_t + Bu_{t+1} + v_{t+1}, \quad (1)$$
$$y_t = Cx_t + w_t, \quad (2)$$
$$x_0 = Bu_0 + v_0, \quad (3)$$

for the physical system, and

$$Pr(s_{t+1}|s_t) = s_{t+1}' \Pi s_t, \quad (4)$$
$$u_t = Ds_t + r_t, \quad (5)$$
$$Pr(s_0) = \pi_0, \quad (6)$$

for the driving actions. The meaning of the variables is as follows: $x_t \in \Re^N$ denotes the hidden state of the LDS, $u_t$ is an input to this system, $v_t$ is the state noise process. Similarly, $y_t \in \Re^M$ is the observed measurement and $w_t$ is the measurement noise. Parameters $A$, $B$ and $C$ are the typical LDS parameters: the state transition matrix, the input matrix and the observation matrix, respectively. The action generator is modeled by a HMM. State variables of this model are written as $s_t$. They belong to the set of $S$ discrete symbols $\{e_0, \ldots, e_{S-1}\}$, where $e_i$ is the unit vector of dimension $S$ with a non-zero element in the $i$-th position. The HMM is defined with the state transition matrix $\Pi$ whose elements are $\Pi(i,j) = Pr(s_{t+1} = e_i|s_t = e_j)$, observation matrix $D$, and an initial state distribution $\pi_0$. The HMM observation noise process is denoted by $r_t$. Note that *the input to the LDS ,u, is the output of the action HMM*.

The mixed state space representation is equivalently depicted by the dependency graph in Figure 2 and can be written as the *joint distribution P*:

$$P(\mathcal{Y}, \mathcal{X}, \mathcal{U}, \mathcal{S}) = Pr(s_0) \prod_{t=1}^{T-1} Pr(s_t|s_{t-1}) \prod_{t=0}^{T-1} Pr(u_t|s_t)$$
$$Pr(x_0|u_0) \prod_{t=1}^{T-1} Pr(x_t|x_{t-1}, u_t) \prod_{t=0}^{T-1} Pr(y_t|x_t), (7)$$

where $\mathcal{Y}, \mathcal{X}, \mathcal{U}$, and $\mathcal{S}$ denote the sequences (with length $T$) of observations and hidden state variables. For instance, $\mathcal{Y} = \{y_0, \ldots, y_{T-1}\}$. Terms $v_t$ and $w_t$ in the physical system formulation are used to denote random noise. We can write an equivalent representation of the physical system in the probability space assuming that the following conditional pdfs are defined:

$$Pr(x_{t+1}|x_t, u_{t+1}) = P_x(x_{t+1} - Ax_t - Bu_{t+1}) \quad (8)$$
$$Pr(y_t|x_t) = P_y(y_t - Cx_t), \quad (9)$$

where $P_x$ and $P_y$ are known, parametric or non-parametric, pdfs. Similarly, the observation pdf of the HMM can be written:

$$Pr(u_t|s_t) = P_u(u_t - Ds_t). \quad (10)$$

Throughout the rest of this paper we assume without loss of generality that the state noise $v$ of the physical system is zero w.p.1 because the HMM observation noise $r_t$ can account for it. The observation noise processes of both the physical system and the HMM are modeled as i.i.d. zero-mean Gaussian:

$$w_t \sim \mathcal{N}(0, R),$$
$$r_t \sim \mathcal{N}(0, Q).$$

Also, assume $B$ to be identity, $B = I$. Input variable $u_t$ can be eliminated from Equations 1 and 5 as an auxiliary variable. Given the above assumptions, the joint pdf of the mixed-state DBN of duration T (or, equivalently, its Hamiltonian[1]) can be written as in Equation 11.

### 2.1   Hidden state inference

The goal of inference in mixed-state DBNs is to estimate the posterior probability of the hidden states of the system ($s_t$ and $x_t$) given some known sequence of observations $\mathcal{Y}$ and the known model parameters. Namely, we need to find the posterior $P(\mathcal{X}, \mathcal{S}|\mathcal{Y}) = Pr(\mathcal{X}, \mathcal{S}|\mathcal{Y})$. In fact, it suffices to find the *sufficient statistics* [5] of the posterior. Given the form of $P$ it is easy to show that these statistics are $\langle [x_t s_t] \rangle$, $\langle [x_t s_t][x_t s_t]' \rangle$, and $\langle [x_t s_t][x_{t-1} s_{t-1}]' \rangle$. The operator $\langle \cdot \rangle$ denotes conditional expectation with respect to the posterior distribution, e.g. $\langle x_t \rangle = \sum_{\mathcal{S}} \int_{\mathcal{X}} x_t P(\mathcal{X}, \mathcal{S}|\mathcal{Y})$.

If there were no action dynamics, the inference would be straightforward – we could infer $\mathcal{X}$ from $\mathcal{Y}$ using LDS inference (RTS smoothing [23]). However, the presence of action dynamics embedded in matrix $\Pi$ makes exact inference more complicated. To see that, assume that the initial distribution of $x_0$ at $t = 0$ is Gaussian, at $t = 1$ the pdf of the physical system state $x_1$ becomes a mixture of $S$ Gaussian pdfs since we need to marginalize over $S$ possible but unknown input levels. At time $t$ we will have a mixture of $S^t$ Gaussians, which is clearly intractable for even moderate sequence lengths. So, it is more plausible to look for an approximate, yet tractable, solution to the inference problem.

---

[1]Hamiltonian $H(x)$ of a distribution $P(x)$ is defined as any positive function such that $P(x) = \frac{\exp(-H(x))}{\sum_\psi \exp(-H(\psi))}$.

$$H(\mathcal{X},\mathcal{S},\mathcal{Y}) = \frac{1}{2}\sum_{t=1}^{T-1}(x_t - Ax_{t-1} - Ds_t)' Q^{-1}(x_t - Ax_{t-1} - Ds_t) + \frac{1}{2}(x_0 - Ds_0)' Q^{-1}(x_0 - Ds_0)$$
$$+ \frac{T}{2}\log|Q| + \frac{NT}{2}\log 2\pi + \frac{1}{2}\sum_{t=0}^{T-1}(y_t - Cx_t)' R^{-1}(y_t - Cx_t) + \frac{T}{2}\log|R| + \frac{MT}{2}\log 2\pi$$
$$+ \sum_{t=1}^{T-1} s_t'(-\log \Pi)s_{t-1} + s_0'(-\log \pi_0) \quad (11)$$

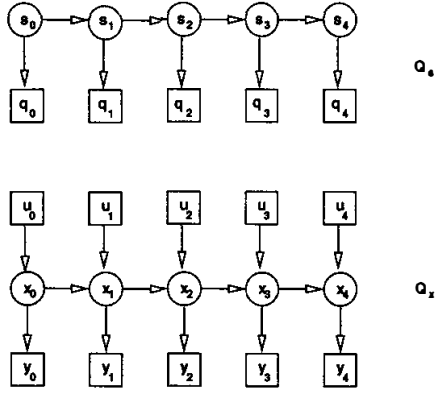

Figure 3: Factorization of the original mixed-state DBN. Factorization reduces the coupled network into a decoupled pair of a HMM ($Q_s$) and a LDS ($Q_x$).

## 2.2 Approximate inference using structured variational inference

Structured variational inference techniques [15] consider a parameterized distribution which is in some sense close to the desired conditional distribution, but is easier to compute. Namely, for a given set of observations $\mathcal{Y}$, a distribution $Q(\mathcal{X},\mathcal{S}|\eta,\mathcal{Y})$ with an additional set of *variational parameters* $\eta$ is defined such that Kullback–Leibler divergence between $Q(\mathcal{X},\mathcal{S}|\eta,\mathcal{Y})$ and $P(\mathcal{X},\mathcal{S}|\mathcal{Y})$ is minimized with respect to $\eta$:

$$\eta^* = \arg\min_{\eta} \sum_{\mathcal{S}} \int_{\mathcal{X}} Q(\mathcal{X},\mathcal{S}|\eta,\mathcal{Y}) \log \frac{P(\mathcal{X},\mathcal{S}|\mathcal{Y})}{Q(\mathcal{X},\mathcal{S}|\eta,\mathcal{Y})}. \quad (12)$$

The dependency structure of $Q$ is chosen such that it closely resembles the dependency structure of the original distribution $P$. However, unlike $P$ the dependency structure of $Q$ *must* allow a computationally efficient inference. In our case we decouple the HMM and LDS as indicated in Figure 3. The two subgraphs of the original network are a HMM $Q_S$ with variational parameters $\{q_0,\ldots,q_{T-1}\} \in \Re^S$ and a LDS $Q_X$ with variational parameters $\{u_0,\ldots,u_{T-1}\} \in \Re^N$. More precisely, the Hamiltonian of the approximating network is defined in Equation 13. The subgraphs are *decoupled*, thus allowing for independent inference, $Q(\mathcal{X},\mathcal{S}|\eta,\mathcal{Y}) = Q_X(\mathcal{X}|\eta,\mathcal{Y})Q_S(\mathcal{S}|\eta)$. This is also reflected in the sufficient statistics of the posterior defined by the approximating network. They are $\langle s_t \rangle$, $\langle s_t s_{t-1}'\rangle$ for the HMM subgraph, and $\langle x_t \rangle$, $\langle x_t x_t'\rangle$, and $\langle x_t x_{t-1}'\rangle$ for the LDS. This, in turn, means that the approximating LDS subnet defines a unimodal posterior.

The optimal values of the variational parameters $\eta = \{q_0,\ldots,q_{T-1}, u_0,\ldots,u_{T-1}\}$ can be obtained by setting the derivative of the KL-divergence w.r.t. $\eta$ to zero. Alternatively, one can employ the theorem of Ghahramani [10] to arrive at the following optimal variational parameters:

$$u_t^* = D\langle s_t \rangle \quad (14)$$
$$q_t^*(i) = e^{d_i' Q^{-1}(\langle x_t \rangle - A\langle x_{t-1}\rangle - \frac{1}{2}d_i)} \quad (15)$$

where $d_i$ denotes the $i$-th column of $D$ and $\langle x_{-1}\rangle \triangleq 0$. To obtain the expectation terms $\langle s_t \rangle = Pr(s_t|q_0,\ldots,q_{T-1})$ we use the inference in the HMM [22] with output "probabilities" $q_t$. Similarly, to obtain $\langle x_t \rangle = E[x_t|u_0,\ldots,u_{T-1},y_0,\ldots,y_{T-1}]$ we perform LDS inference (Rauch-Tung-Streibel smoothing [23]) with inputs $u_t$. Since $u_t$ in subgraph $Q_X$ depends on $\langle s_t \rangle$ from subgraph $Q_S$ and $q_t$ depends on $\langle x_t \rangle$, Equations 14 and 15 together with the inference solutions form a set of *fixed-point equations*. Solution of this fixed-point set yields a tractable approximation to the intractable original posterior. Error bounds of the approximation are easy to derive and can be found in [20].

The variational inference algorithm for mixed-state DBNs can now be summarized as:

```
error = ∞;
Initialize ⟨x⟩;
while (error > maxError) {
    Find q_t from ⟨x_t⟩ using Equation 15;
    Estimate ⟨s_t⟩ from q_t using HMM inference;
    Find u_t from ⟨s_t⟩ using Equation 14;
    Estimate ⟨x_t⟩ from y_t and u_t using LDS
        inference;
    Update approximation error;
}
```

$$H_Q(\mathcal{X},\mathcal{S},\mathcal{Y}) = \frac{1}{2}\sum_{t=1}^{T-1}(x_t - Ax_{t-1} - u_t)'Q^{-1}(x_t - Ax_{t-1} - u_t) + \frac{1}{2}(x_0 - u_0)'Q^{-1}(x_0 - u_0)$$
$$+ \frac{T}{2}\log|Q| + \frac{NT}{2}\log 2\pi + \frac{1}{2}\sum_{t=0}^{T-1}(y_t - Cx_t)'R^{-1}(y_t - Cx_t) + \frac{T}{2}\log|R| + \frac{MT}{2}\log 2\pi$$
$$+ \sum_{t=1}^{T-1} s_t'(-\log\Pi)s_{t-1} + s_0'(-\log\pi_0) + \sum_{t=0}^{T-1} s_t'(-\log q_t) \qquad (13)$$

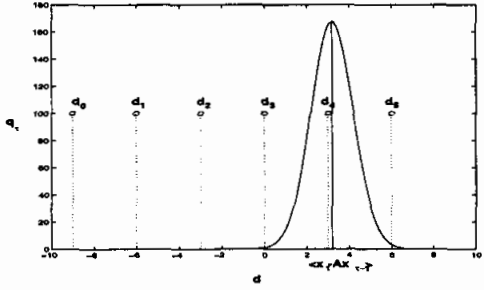

Figure 4: Variational parameter $q_t$ as a function of input level $d$. Shown is a set of six different input levels $d_0$ through $d_5$. The function attains maximum for input level $d_4$ which is closest to the global maximum at $d^* = \langle x_t \rangle - A \langle x_{t-1} \rangle$.

From Equation 14 and the factorization of the network defined in Equation 13 it is evident that $u_t$ can be viewed as the estimated input of the LDS, based on the estimates of the hidden states of the HMM subgraph. The input at time $t$ is estimated to be a linear combination of all possible inputs $d_i$ weighted by their corresponding probabilities $\langle s_t(i) \rangle$, $D \langle s_t \rangle = \sum_{i=0}^{N-1} d_i \langle s_t(i) \rangle$.

The meaning of $q_t$ is not immediately obvious. Based on Equation 13, $q_t$ can be viewed as the probabilities of some fictional discrete-valued inputs presented to the HMM subgraph. These probabilities are related to the estimates of the states $x_t$ of the LDS through Equation 15. To better understand the meaning of this dependency consider the plot in Figure 4 of $q_t$ versus $d = d_i$ for a fixed value of the difference $\langle x_t \rangle - A \langle x_{t-1} \rangle$ and unit variance $Q$. Clearly, the function assumes a maximum value for $d = \langle x_t \rangle - A \langle x_{t-1} \rangle$. If we have a set of discrete values of $d$ corresponding to $N$ possible LDS input levels $d_i$, $q_t(i)$ would be maximized for $d_i$ closest to the estimated difference $\langle x_t \rangle - A \langle x_{t-1} \rangle = \langle u_t \rangle$. Thus, those states of the HMM are favored which produce inputs "closer" to the ones estimated from the LDS dynamics.

## 3 Variational learning in mixed-state DBNs

Learning in mixed-state DBNs can be formulated as the problem of ML learning in general Bayesian networks. It was shown in [15] that structured variational inference can be viewed as the *expectation* step of a generalized EM algorithm [13, 19]. The maximization step then yields

$$\theta^* = \arg\max_\theta \sum_\mathcal{S} \int_\mathcal{X} Q(\mathcal{X},\mathcal{S}|\mathcal{Y},\eta^*)\log P(\mathcal{X},\mathcal{S},\mathcal{Y}),$$

where $\theta$ is the set of parameters of pdf $P$. In our case, the parameters are $\{A,C,D,Q,R,\Pi,\pi_0\}$.

Given the sufficient statistics obtained in the inference phase, it is easy to show that the following *parameter update equations* result from the Maximization step:

$$A^* = \left(\sum_{t=1}^{T-1}\langle x_t x_{t-1}'\rangle - D^*\langle s_t\rangle\langle x_{t-1}'\rangle\right)$$
$$\left(\sum_{t=1}^{T-1}\langle x_{t-1} x_{t-1}'\rangle\right)^{-1}$$
$$Q^* = \frac{1}{T}\sum_{t=0}^{T-1}\langle x_t x_t'\rangle - A^*\langle x_{t-1}x_t'\rangle - D^*\langle s_t\rangle\langle x_t'\rangle$$
$$D^* = \left(\sum_{t=0}^{T-1}\langle x_t\rangle\langle s_t'\rangle - A^*\langle x_{t-1}\rangle\langle s_t'\rangle\right)$$
$$\left(\sum_{t=0}^{T-1}\langle s_t s_t'\rangle\right)^{-1}$$
$$C^* = \left(\sum_{t=0}^{T-1} y_t \langle x_t'\rangle\right)\left(\sum_{t=0}^{T-1}\langle x_t x_t'\rangle\right)^{-1}$$
$$R^* = \frac{1}{T}\sum_{t=0}^{T-1}(y_t y_t^t - C^*\langle x_t\rangle y_t')$$
$$\Pi^* = \left(\sum_{t=1}^{T-1}\langle s_t s_{t-1}'\rangle\right)\left(\sum_{t=1}^{T-1}\langle s_t s_t'\rangle\right)^{-1}$$
$$\pi_0^* = \langle s_0\rangle.$$

All the variable statistics are evaluated before updating any parameters. Notice that the above equations represent a generalization of the parameter update equations of zero-input LDS models [9].

## 4 Variational v.s. greedy approximation

As stated before, structured variational approximation has an advantage over several greedy or Monte Carlo approaches in that it consistently decreases the relative entropy between the approximating distribution and the true posterior distribution. To demonstrate this advantage we consider the following simple case.

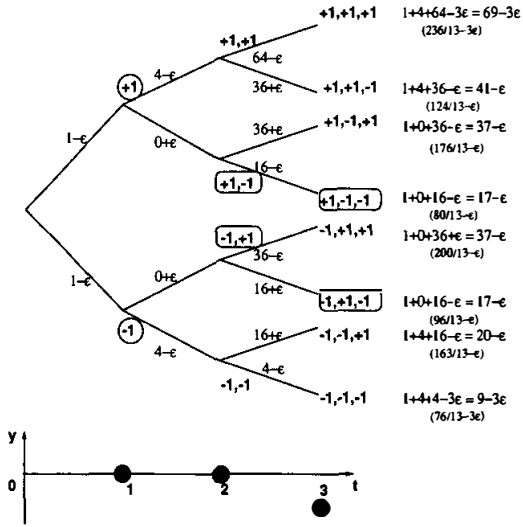

Figure 5: Trellis for 8 different input sequences. Nodes indicate sequences of LDS inputs and arcs indicate transition costs $(y_t - x_{t|t} - u_t)^2/Var_t - \log(P(s_t = i|s_t = j))$. Final costs for $k = 0$ and $k = 1$ are shown in the rightmost column. Costs in parentheses are for $k = 1, R = .5$.

Let the LDS parameter of the model be given as: $A = 1, C = 1, Q = kR, k \geq 0, R > 0, x_0 = 0$ Furthermore, let the HMM parameters be

$$\Pi = \begin{bmatrix} 0.5 + \epsilon & 0.5 - \epsilon \\ 0.5 - \epsilon & 0.5 + \epsilon \end{bmatrix}, \pi_0 = \begin{bmatrix} 0.5 \\ 0.5 \end{bmatrix},$$

where $\epsilon \geq 0$. Finally, assume that each discrete state can produce an input to the LDS of $D = [-1 1]$.

Consider the case where a sequence of discrete states $S_T = \{1, 1, 1\}$ (LDS input sequence $U_T = \{-1-1-1\}$) resulted in a sequence of noisy measurements $Y_T = \{0, 0, -5\}$. For a moment, assume that $k = 0$, i.e., there is no LDS state noise. Then, the trellis in Figure 5 summarizes costs for all possible discrete state paths[2]

The greedy (truncated Viterbi) approximation selects one best path based on the minimal partial cost of the LDS innovation and HMM transition (arcs in Figure 5.) Let's further simplify the analysis by assuming that $\epsilon = 0$. Hence, at $t = 1$ either $+1$ or $-1$ sequence can be selected with equal cost. Assuming that $-1$ was selected at $t = 1$, at time $t = 2$ $+1$ is selected as it yields the minimal partial cost $(0 + 16.)$ Finally, at $t = 3$ discrete input $-1$ is selected again as the sequence $-1, +1, -1$ yields the least cost of 17. Clearly, this greedy solution is suboptimal! In fact, it is easy to see that the optimal input sequence $-1, -1, -1$ results in the total cost of 9. Similar analysis (with somewhat different, non-integer, costs) and conclusions hold for non-zero LDS state noise $k > 0$.

---
[2]We assume that $\epsilon$ is small, hence the HMM state transition cost is $-\log(1 \pm \epsilon) = \mp\epsilon$.

Variational approximation, on the other hand, does not suffer from this problem. Table 1 shows that for $k = 1, R = 0.5$ and a particular initial condition $(q_t(i = 1) = q_t(i = -1) = 0, \forall t)$ variational inference yields a discrete state sequence $-1, -1, -1$. In fact, any initial conditions yields the generating discrete state sequence.

Table 1: Optimal variational parameters for $k = 1, R = 0.5, \epsilon = 0$.

| Iter. | $q_1(-1)$ | $q_2(-1)$ | $q_3(-1)$ | $u_1$ | $u_2$ | $u_3$ |
|---|---|---|---|---|---|---|
| 1 | 0.000 | 0.000 | 0.000 | 0.683 | 0.823 | 0.979 |
| 2 | -0.367 | -0.646 | -0.958 | 0.647 | 0.854 | 0.991 |
| 3 | -0.293 | -0.709 | -0.983 | 0.626 | 0.865 | 0.991 |
| 4 | -0.252 | -0.731 | -0.983 | 0.616 | 0.870 | 0.991 |
| 5 | -0.233 | -0.741 | -0.983 | 0.611 | 0.872 | 0.991 |
| 6 | -0.223 | -0.745 | -0.983 | 0.609 | 0.873 | 0.991 |
| 7 | -0.219 | -0.747 | -0.983 | 0.608 | 0.874 | 0.991 |
| 8 | -0.217 | -0.748 | -0.983 | 0.608 | 0.874 | 0.991 |
| 9 | -0.216 | -0.749 | -0.983 | 0.607 | 0.874 | 0.991 |
| 10 | -0.215 | -0.749 | -0.983 | 0.607 | 0.874 | 0.991 |

We note, however, that the presence of "non-uniform" HMM parameters (i.e., $\epsilon > 0$.) may influence the results of our analysis. For instance, in the case when $k = 0$ for $\epsilon > 2$ the greedy truncated Viterbi approximation does in fact achieve the optimal cost as it select input $-1$ at $t = 2$. The effect of "non-uniform" HMM is even more obvious for $k > 0$, as the necessary threshold for $\epsilon$ reduces.

Hence, structured variational approximation stands out in the cases where the HMM only slightly differentiates between possible discrete state sequences and the variance of the LDS noise processes are sufficiently small.

## 5 Analysis and recognition of hand gestures acquired by a computer mouse

To demonstrate feasibility of the mixed-state DBN framework we consider the task of classifying a set of symbols drawn using a computer mouse. We defined four classes of symbols: arrow, erase, circle, and wiggle (see Figure 6.) The task in question was to model each of the four symbols with a combination of LDS and HMM. The LDS part modeled the Newtonian dynamics of the mouse motion. Namely, we assumed that the mouse motion can be modeled as a planar motion of a point-mass particle with piece-wise constant acceleration:

$$\frac{d\,position(t)}{dt} = velocity(t)$$
$$\frac{d\,velocity(t)}{dt} = u(t) + noise(t).$$

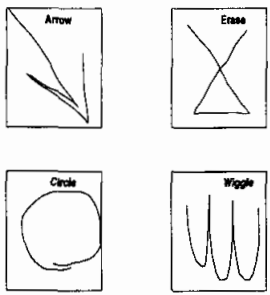

Figure 6: Examples of four symbols produced by computer mouse motion.

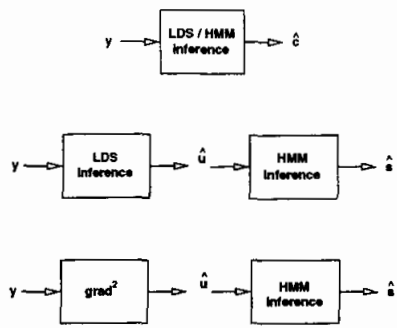

Figure 7: Three ways of modeling mouse acquired symbols. From top to bottom: completely coupled LDS and HMM (mixed-state HMM), decoupled adapted LDS and HMM, and decoupled fixed LDS and HMM.

This leads to a discrete-time LDS with known $A$, $B$ and $C$ and unknown $Q$ and $R$ (cf [1]). On the other hand the HMM models the driving force (action) that causes the motion. The mixed-state DBN model is contrasted with two decoupled model

- Decoupled adapted LDS and HMM. Namely, the LDS is adapted to "best" model the dynamics of the mouse motion of each symbol when the driving force $u_t$ is assumed to be quasi-constant with additive white noise, $u_t = u_{t-1} + n_{u,t}$. The HMM is consequently employed to model the quasi-constant driving force $\langle u_t \rangle$ inferred by the LDS.

- Decoupled fixed LDS and HMM. In this case, the LDS is assumed to be fixed for all four symbols. In particular, we estimated the driving force using numerical gradient approximation: $u_t = grad(grad(x_t))$, where $grad(x_t) = \frac{x_{t+1} - x_{t-1}}{2 \cdot \Delta T}$. Again, an HMM is used to model the estimated driving force.

All three model classes are depicted in Figure 7.

For each of the three models the same action state spaces are assumed. The number of action states is proportional to the number of strokes necessary to produce each symbol. Thus, the action model of the arrow symbol had eight states (two times four strokes), erase has six states, circle four states, and wiggle six. Furthermore, each symbol's state transitions are limited to left-to-right: from current state the action can only transition back to itself or to only one other not-yet-visited state. In the two decoupled symbol models, we model the observations $u_t$ of the action models as variable mean Gaussian processes with identical variances at every action state[3]. Model parameters are learned from data using the ML learning framework.

The data set consists of 136 examples of each symbol (a total of 4 × 136 examples). Symbols were acquired from normalized[4] mouse movements sampled at $\Delta T = 100ms$ intervals. To test the models' performance we used rotation error counting (cross-validation) method with four rotational sets [7]. For each test sample and each symbol model the likelihood of the sample was appropriately obtained. For instance, in the case of symbols modeled by mixed-state DBNs, variational inference with a relative error threshold of $10^{-3}$ was used to estimate the lower bound on likelihood. One example of mixed-state DBN-based decoding of the "arrow" symbol is shown in Figure 8. For the fixed LDS and gradient-based LDS/HMM models, likelihood was obtained using the standard HMM and LDS inference.

Classification test were performed on two sets of data: noise-free and noisy. Classification of noisy symbols is of particular interest since it introduces variability that may pose a challenge to decoupled classification models. The noisy data set was constructed by adding i.i.d. zero mean Gaussian noise with standard deviation of 0.01 to noise-free examples (see Figure 9). Models of the four symbols trained on noise-free samples were now tested on the noisy data. Classification results are summarized in Table 2 and Figure 10.

---

[3]Even though it is a usual practice to allow the variance to vary from action state to action state, for sake of compatibility with the fixed variance mixed-state HMM we decided to keep the other HMMs' observation variances fixed.

[4]Symbol were scaled to $[0, 1] \times [0, 1]$ unit area and directionally aligned.

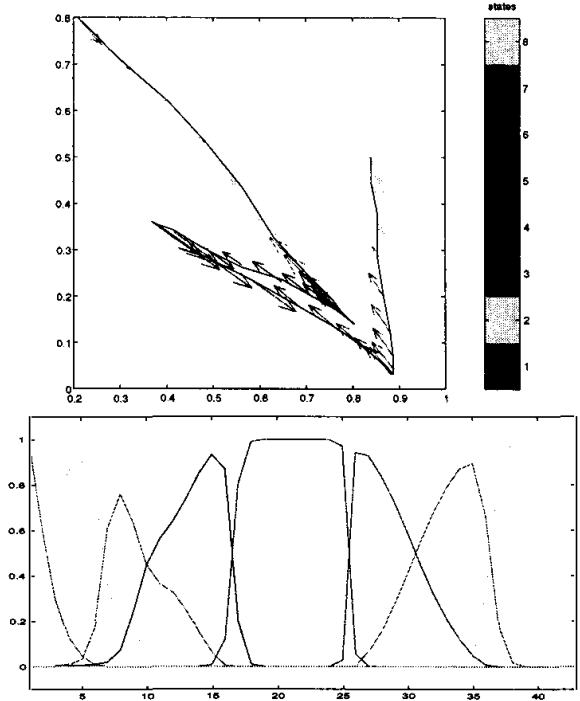

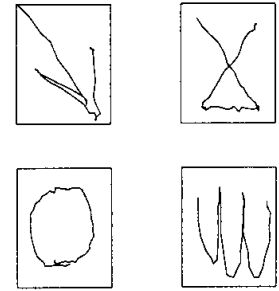

Figure 9: Samples of four mouse acquired symbols corrupted by additive zero mean white noise with standard deviation of 0.01.

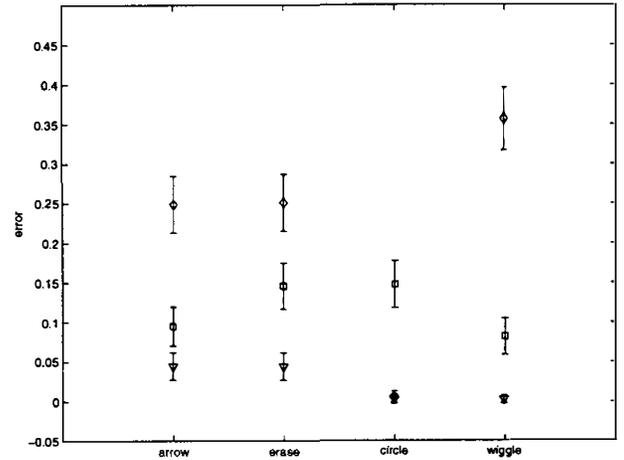

Figure 8: Estimates of action states for "arrow" symbol. Top graph depicts the symbol and the estimated driving force. Bottom graph shows estimates of action states obtained using variational inference. Colors in both graphs indicate action states.

Table 2 and Figure 10 in this case indicate that with 95% confidence completely coupled mixed-state HMM models had significantly better performance that both fixed and adapted decoupled LDS/HMM classifiers (with the exception of mixed-state and fixed LDS "circle" models). Of course, the tradeoff is as always in increased computational complexity of the mixed-state models. We note, however, that on the average the iterative scheme of the mixed-state models required only about 5 to 10 iterations to converge.

| Model | Arrow | Erase | Circle | Wiggle |
|---|---|---|---|---|
| mixed-state HMM | 4.36 (0.89) | 4.36 (0.89) | 0.18 (0.26) | 0.18 (0.26) |
| gradient fixed LDS/HMM | 9.45 (1.25) | 14.55 (1.51) | 14.73 (1.51) | 8.18 (1.18) |
| decoupled adapted LDS/HMM | 24.91 (1.84) | 25.09 (1.85) | 0.55 (0.36) | 35.64 (2.04) |

Table 2: Error estimates [%] and error estimate variances ([%]) for noisy mouse symbol classification.

Figure 10: Classification error estimates of four noisy mouse symbols. Show are 95% confidence intervals for error counts. The coupled mixed-state DBN performs significantly better than the decoupled adapted and fixed LDS/HMM models. (▽ - mixed-state DBN, □ - fixed (gradient) LDS/HMM, ◇ - adapted LDS/HMM.)

## 6  Summary

We presented a novel "mixed-state" dynamic graphical model for modeling of time-series that *fuses* the typical models of driving actions (HMMs) with continuous state models of physical systems (LDSs) and we introduced a structured variational technique for inference and learning in the otherwise intractable model. The structured variational technique yields a best unimodal approximation to the exact polymodal posterior. For the task of classifying patterns drawn using a mouse, we found that the model performed better than a greedy Viterbi-based algorithm.

### Acknowledgments

This work was supported in part by National Science Foundation grant IRI-96-34618 and in part by the Army Research Laboratory Cooperative Agree-

ment No. DAAL01-96-2-0003.